\documentclass{article}

\usepackage{microtype}
\usepackage{graphicx}
\usepackage{subfigure}
\usepackage{booktabs} 

\usepackage{hyperref}

\usepackage[ruled]{algorithm2e}

\usepackage{inconsolata}
\usepackage{subfigure}
\usepackage{amssymb,amsthm,dsfont,mathrsfs}
\usepackage{multirow}
\usepackage{mathrsfs}
\usepackage{float}
\usepackage{hyperref}
\usepackage{adjustbox}
\usepackage{booktabs}
\usepackage{multirow}
\usepackage{lscape}
\usepackage{xcolor}
\usepackage{colortbl}
\usepackage{xspace}
\usepackage{tabularx}
\usepackage[most]{tcolorbox}
\usepackage{bbding} 
\usepackage{arydshln}
\usepackage[frozencache,cachedir=.]{minted} 
\usepackage{graphicx}
\usepackage{shadowtext}
\definecolor{lightblue}{RGB}{210, 220, 250}
\newcommand{\blue}{\cellcolor{lightblue}}

\usepackage[accepted]{icml2024}

\usepackage{amsmath}
\usepackage{amssymb}
\usepackage{mathtools}
\usepackage{amsthm}

\usepackage[capitalize,noabbrev]{cleveref}

\theoremstyle{plain}

\theoremstyle{definition}

\theoremstyle{remark}

\usepackage[textsize=tiny]{todonotes}

\icmltitlerunning{Training Large Language Models for Reasoning through Reverse Curriculum Reinforcement Learning}

\begin{document}

\twocolumn[
\icmltitle{Training Large Language Models for Reasoning 
 \texorpdfstring{\\}{} through Reverse Curriculum Reinforcement Learning}
\icmlsetsymbol{equal}{*}
\icmlsetsymbol{correpsonding}{$\dagger$}

\begin{icmlauthorlist}
\icmlauthor{Zhiheng Xi}{Fudan CS,equal,correpsonding}
\icmlauthor{Wenxiang Chen}{Fudan CS,equal}
\icmlauthor{Boyang Hong}{Fudan CS,equal}
\icmlauthor{Senjie Jin}{Fudan CS,equal}
\icmlauthor{Rui Zheng}{Fudan CS,correpsonding}
\icmlauthor{Wei He}{Fudan CS}
\icmlauthor{Yiwen Ding}{Fudan CS}
\icmlauthor{Shichun Liu}{Fudan CS}
\icmlauthor{Xin Guo}{Fudan CS}
\icmlauthor{Junzhe Wang}{Fudan CS}
\icmlauthor{Honglin Guo}{Fudan CS}
\icmlauthor{Wei Shen}{Fudan CS}
\icmlauthor{Xiaoran Fan}{Fudan CS}
\icmlauthor{Yuhao Zhou}{Fudan CS}
\icmlauthor{Shihan Dou}{Fudan CS}
\icmlauthor{Xiao Wang}{Fudan CS}
\icmlauthor{Xinbo Zhang}{BDR}
\icmlauthor{Peng Sun}{BDR}
\icmlauthor{Tao Gui}{Fudan IMLL,correpsonding}
\icmlauthor{Qi Zhang}{Fudan CS,correpsonding}
\icmlauthor{Xuanjing Huang}{Fudan CS}
\end{icmlauthorlist}

\center
{\tt zhxi22@m.fudan.edu.cn},
{\tt \{rzheng20,tgui,qz\}@fudan.edu.cn}

\icmlaffiliation{Fudan CS}{School of Computer Science, Fudan University}
\icmlaffiliation{Fudan IMLL}{Institute of Modern Languages and Linguistics, Fudan University}
\icmlaffiliation{BDR}{ByteDance Research}

\icmlkeywords{Machine Learning, ICML}

\vskip 0.3in
]




\printAffiliationsAndNotice{\icmlEqualContribution}

\begin{abstract}
In this paper, we propose \textbf{R}$^3$: Learning \textbf{R}easoning through \textbf{R}everse Curriculum \textbf{R}einforcement Learning (RL), a novel method that employs only outcome supervision to achieve the benefits of process supervision for large language models. The core challenge in applying RL to complex reasoning is to identify a sequence of actions that result in positive rewards and provide appropriate supervision for optimization. Outcome supervision provides sparse rewards for final results without identifying error locations, whereas process supervision offers step-wise rewards but requires extensive manual annotation. \textbf{R}$^3$ overcomes these limitations by learning from correct demonstrations. Specifically, \textbf{R}$^3$ progressively slides the start state of reasoning from a demonstration's end to its beginning, facilitating easier model exploration at all stages. Thus, \textbf{R}$^3$ establishes a step-wise curriculum, allowing outcome supervision to offer step-level signals and precisely pinpoint errors. Using Llama2-7B, our method surpasses RL baseline on eight reasoning tasks by $4.1$ points on average. Notebaly, in program-based reasoning on GSM8K, it exceeds the baseline by $4.2$ points across three backbone models, and without any extra data, Codellama-7B + \textbf{R}$^3$ performs comparable to larger models or closed-source models.\footnote{Our codes and data are available at Github : \href{https://github.com/WooooDyy/LLM-Reverse-Curriculum-RL}{https://github.com/WooooDyy/LLM-Reverse-Curriculum-RL}.}
\end{abstract}

\section{Introduction}

\begin{figure*}[t]
    \includegraphics[width=0.85\linewidth]{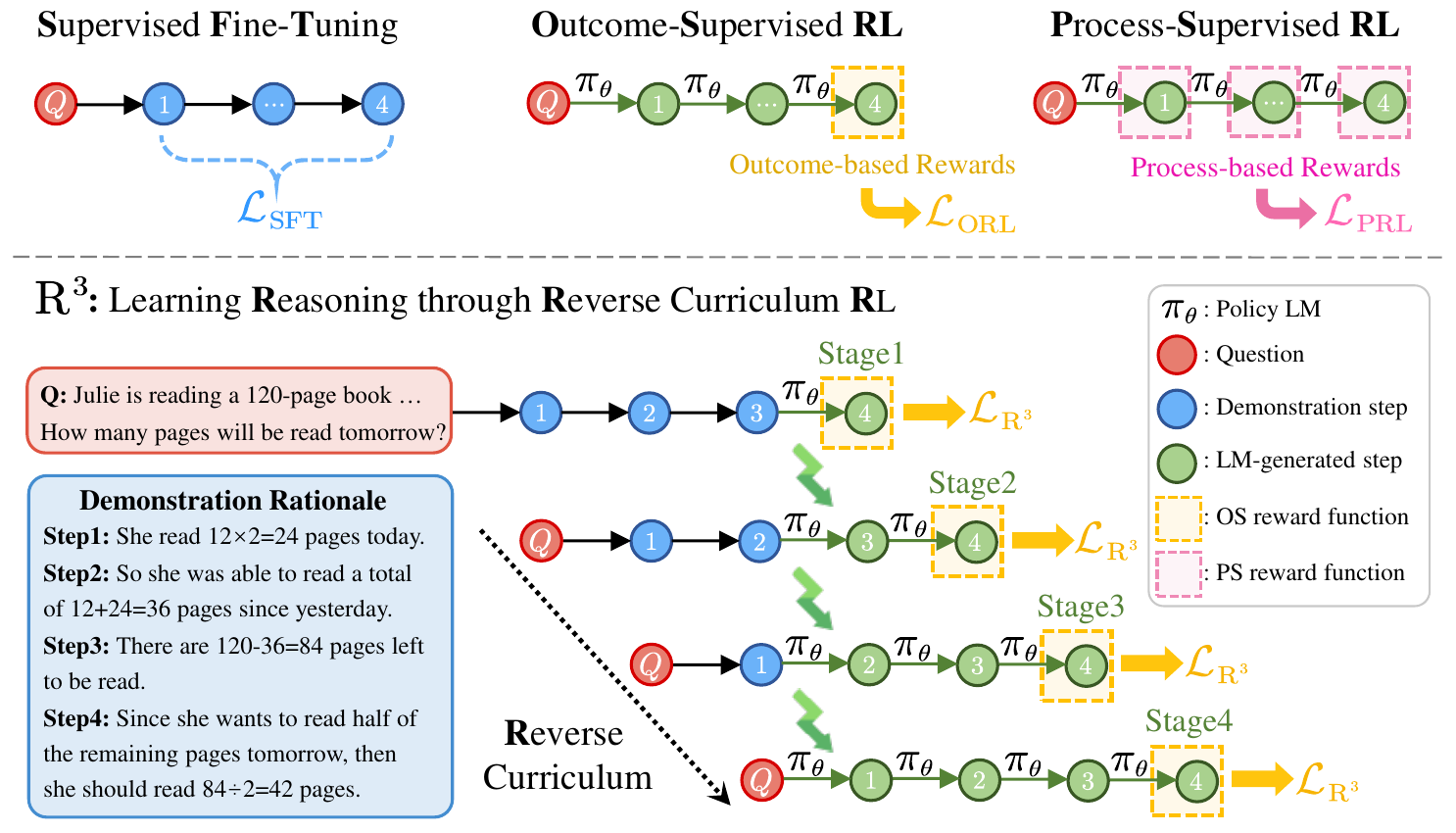}
    \centering
    \vspace{-0.3cm}
 	\caption{
  Schematic comparison between R$^3$ and other methods for training LLMs for reasoning. 
  $\mathcal{L}_{(\cdot)}$ represents the optimization objective for each method.
  Supervised Fine-Tuning optimizes models using annotated rationales, without additional exploration. 
  In RL, the model first generates a reasoning path and receives supervisory signals for optimization.
  Outcome-Supervised (OS) RL rewards the final result, while Process-Supervised (PS) RL rewards each reasoning step. 
  The proposed R$^3$ provides approximately step-by-step supervisory signals similar to PS with only an OS reward function. 
  }\label{fig:main}
  \vspace{-0.1cm}

\end{figure*}

Large language models (LLMs) have made impressive advancements in complex, multi-step reasoning, by prompting or learning to generate solutions in a step-by-step Chain-of-Thought manner \citep{DBLP:conf/nips/Wei0SBIXCLZ22,DBLP:conf/nips/KojimaGRMI22,DBLP:conf/emnlp/KimJKJYSS23}.
Training a language model specialized in reasoning is proved to be superior to prompting-based approaches \citep{DBLP:journals/corr/abs-2211-14275, DBLP:journals/corr/abs-2309-12284}.
However, Supervised Fine-tuning (SFT) focuses on imitating human demonstrations, requiring large-scale, diverse annotations to achieve generalization \cite{DBLP:journals/corr/abs-2305-20050,DBLP:journals/corr/abs-2308-01825,DBLP:conf/emnlp/ShenYLSJ0021}.
Reinforcement learning (RL) offers a viable alternative to improve reasoning via exploration and learning \cite{DBLP:journals/corr/abs-2204-05862,DBLP:conf/nips/Ouyang0JAWMZASR22,DBLP:journals/corr/abs-2307-04964,DBLP:journals/corr/abs-2308-09583}.

\begin{table}[t]
\centering
\caption{Comparison of the proposed R$^3$ with other supervision methods in terms of three key features. 
\textbf{Golden} means whether the supervisory signals are based on golden labels (e.g., correctness) or human preference; 
\textbf{Human-Annotation-free} indicates that the method does not require detailed annotations for each intermediate step; \textbf{Step-level Sup.} means whether the method can provide step-by-step supervisory signals.
}
\resizebox{0.48\textwidth}{!}{
\begin{tabular}{lccc}
\toprule

 \multirow{3}{*}{\textsc{Feature}} & \multicolumn{3}{c}{\textsc{Supervision Method}}   \\

 \cmidrule(l){2-4}

 &Outcome Sup. &Process Sup. & Ours\\
\cmidrule(l){1-1} 
\cmidrule(l){2-4}

Golden &\textcolor{cyan}{\CheckmarkBold} &\textcolor{red}{\XSolidBrush}  &\textcolor{cyan}{\CheckmarkBold}  \\
Human-Annotation-free &\textcolor{cyan}{\CheckmarkBold} &\textcolor{red}{\XSolidBrush}  & \textcolor{cyan}{\CheckmarkBold}  \\
Step-level Sup. &\textcolor{red}{\XSolidBrush} &\textcolor{cyan}{\CheckmarkBold}  & \textcolor{cyan}{\CheckmarkBold}  \\

\midrule
\end{tabular}
}

\label{table:different supervision comparison}
\vspace{-1.5mm}
\end{table}

When applying RL to complex reasoning tasks, the core challenge lies in identifying a sequence of actions that yield positive rewards and providing appropriate supervisory signals for optimization \citep{sutton1998introduction}.
On one hand, as task difficulty increases, so does the complexity and length of the reasoning chain.
LLMs struggle with the accumulation of errors and uncertainties across multiple intermediate steps \cite{DBLP:journals/corr/abs-2305-20050,DBLP:journals/corr/abs-2311-09724,zhang2023siren}. 
The increase of reasoning steps leads to an exponential growth in the search space for reasoning, making it challenging to obtain correct final results \cite{xie2023self}.
On the other hand, existing methods for supervised signals require a trade-off between feedback quality and annotation cost \cite{DBLP:journals/corr/abs-2211-14275}.
Outcome supervision (OS, \citealp{DBLP:journals/corr/abs-2110-14168,DBLP:journals/corr/abs-2311-09724}) rewards only the final outcome (top center in Figure \ref{fig:main}), but sparse rewards make it difficult to determine which actions led to success or failure \cite{wang2023math}.
Process supervision (PS, \citealp{DBLP:journals/corr/abs-2211-14275,DBLP:journals/corr/abs-2305-20050}) provides detailed feedback at every step of reasoning (top right in Figure \ref{fig:main}), but this approach requires highly skilled annotators to select better reasoning paths, significantly increasing costs \cite{DBLP:journals/corr/abs-2305-20050}.

In this work, we propose \textbf{R}$^3$: Learning \textbf{R}easoning through \textbf{R}everse Curriculum \textbf{R}einforcement Learning (bottom in Figure \ref{fig:main}) to address the limitations. It employs only outcome supervision to achieve an effect similar to process supervision.
Specifically, R$^3$ let the model begin reasoning from a state sampled from a correct demonstration, and provide feedback to supervise the generated actions with outcome supervision. By slowly moving the start state from the end of the demonstration to the beginning, the model faces an easy exploration problem at each point where it is likely to succeed, since it has already learned to solve most of the remaining parts. In this way, a curriculum of gradually increasing exploration difficulty is created, and we can provide approximately step-by-step supervisory signals for the model.

This method facilitates the model's exploration as it shortens the reasoning chain and narrows the sampling space, aiding the model in gaining positive rewards more efficiently.
We can interpret \textbf{R}$^3$ as a form of dynamic programming \cite{bertsekas2012dynamic}. If $N$ reasoning steps are required to obtain a reward, this reasoning can now be learned in a time that is linear in $N$, rather than exponential \cite{DBLP:conf/corl/FlorensaHWZA17, openai_1}.
To improve the training stability and model generalization, we mix the start states of various exploration difficulties for training.
Thorough experiments on Llama2-7B demonstrate that R$^3$ outperforms both the SFT and RL baselines across eight reasoning tasks, achieving an average improvement of $5.4$ points and $4.1$ points, respectively.
Notably, in program-based reasoning on GSM8K, it surpasses SFT and RL by an average of $11.4$ points and $4.2$ points, respectively. Moreover, Codellama-7B + R$^3$ outshines models that use extra annotated data like MAmmoTH-Coder \cite{DBLP:journals/corr/abs-2309-05653} and Tora \cite{DBLP:journals/corr/abs-2309-17452}, and is comparable to larger or closed-source models such as GPT-3.5-Turbo.

In summary, we make the following contributions:
\begin{enumerate}
    \item We propose {R}$^3$, a novel method which employs outcome supervision to achieve an effect similar to process supervision, to enhance the reasoning ability of LLMs.
    \item We conduct extensive experiments across eight reasoning tasks to highlight the effectiveness of our method. Furthermore, we showcase the superiority of {R}$^3$ in program-based reasoning through its application on three models for solving math problems.
    \item We perform in-depth ablation and analysis to provide insights into the training dynamics of R$^3$ and how it works.
\end{enumerate}

\section{RL with Outcome and Process Supervision} \label{sec:RL with Outcome Supervsion and Process Supervision}
We use RL notations to describe the language generation process. At each timestep $t$, the policy language model (LM) $\pi_\theta^{RL}$ parameterized by $\theta$ receives a state $s_t$, which consists of the input prompt and the generated text up to this point. Then, the policy's action $a_{t+1}$ is to generate next token conditioned on the state, and the probability is as $\pi_\theta (a_{t+1}|{s_{t}})$. After that, the environment returns a reward $r(s_t,a_{t+1})$, and the state is transitioned to $s_{t+1}$ with the transition probability $p(s_{t+1}|s_t,a_{t+1})$. The goal of RL is to find an optimal policy to maximize the cumulative reward (i.e., return) over a trajectory $\tau = \{s_0, a_1,...,s_T,a_T\}$ where $s_0$ is the initial state (i.e., the prompt) and $T$ is the length of actions. The general form of the policy gradient is gaven as \citep{DBLP:conf/icml/MnihBMGLHSK16}:
\begin{equation}
\begin{aligned} \label{eqn: policy gradient}
     \mathbb{E}_{\tau \sim \pi^\mathrm{RL}_\theta} \left[ \sum_{t=1}^T   \nabla_\theta \log \pi^\mathrm{RL}_\theta(a_{t}|s_{t-1}) R(s_{t-1},a_t) \right],
\end{aligned}
\end{equation}
where $\mathbb{E}_{\tau \sim \pi^\mathrm{RL}_\theta}$ refers to the expectation under the distribution of trajectories sampled from the policy $\pi^\mathrm{RL}_\theta$. The return $R(s_{t-1},a_t) =\sum^T_{t'=t} \gamma^{t'-t+1} r(s_{t'-1}, a_{t'})$ is the discounted sum of rewards from timestep $t$ with factor $\gamma \in [0,1)$. 
With this gradient, we can perform gradient ascent to optimize the model.
If the return is favorable, the actions are ``reinforced'' by increasing their probability of being selected. 
Given a dataset $\mathcal{D}=\{(s_0^i,\mathbf{a}^i)\}_{i=1}^N$ of $N$ pairs of input ${s_0}$ and human-generated output sequence $\mathbf{a}$, where $\mathbf{a} = (a_1,a_2,...,a_T) $ and the whole trajectory is $\tau = \{s_0, a_1,...,s_{T-1},a_T\}$. The policy gradient becomes:
\begin{equation}
\begin{aligned} 
  & \mathbb{E}_{{s_0}\sim\mathcal{D}} \Biggl[ \mathbb{E}_{\tau \sim \pi^\mathrm{RL}_\theta(\cdot|s_0)} \Biggl[ \\ 
   & \ \ \ \ \ \ \ \ \ \ \ \ \ \ \ \ \sum_{t=1}^T \nabla_\theta \log  \pi^\mathrm{RL}_\theta(a_{t}|s_{t-1}) R(s_{t-1},a_t) \Biggr] \Biggr].
\end{aligned}
\end{equation}

\subsection{Outcome Supervision and Process Supervision} \label{sec:Outcome Supervision and Process Supervision}
Here we present the operating mechanisms of outcome supervision and process supervision, along with their advantages and limitations, as briefly summarized in Table \ref{table:different supervision comparison}.
\paragraph{Outcome supervision.}
In outcome supervision, only the final result of the sampled sequence is assigned a reward score, and the score for other tokens are $0$ \citep{DBLP:journals/corr/abs-2110-14168,DBLP:journals/corr/abs-2311-09724}:
$$
r_o(s_{t-1},a_{t}) = \left\{  \begin{aligned} 
& {rf}_o(s_{t-1},a_{t}),  \ \ &t = T \\
& 0, \ \ &t \neq T
\end{aligned}
\right. 
$$
where $rf_o(\cdot)$ is a reward function that returns $1$ is the answer is correct else $0$.
In this paradigm, we don't require detailed annotations for each reasoning step or the training a reward model to allocate rewards.
Instead, the golden answer to the question is enough. This supervision is solely based on the correctness, not on the preference of humans. Despite this simplicity, the supervisory signals are sparse, making it challenging for the policy LM to pinpoint reasoning errors accurately. 
The policy may fall into aimless exploration and struggle in obtaining positive rewards due to the large action space of the LM and the long decision-making chain.

\paragraph{Process supervision.} In process supervision, a reward model $rm_p(\cdot)$ is trained to assign a reward score for each intermediate reasoning step \cite{DBLP:journals/corr/abs-2211-14275,DBLP:journals/corr/abs-2305-20050}:
$$
r_p(s_{t-1},a_{t}) = \left\{  \begin{aligned} 
& {rm}_p(s_{t-1},a_{t}),  \ \ &t \in \mathcal{T}^{Delimiter} \\
& 0, \ \ &t \notin \mathcal{T}^{Delimiter}
\end{aligned}
\right. 
$$
where $\mathcal{T}^{Delimiter}$ represents the set of timesteps that delimite each step (e.g., newline or some special symbols). In this paradigm, the rewards are dense, then provide more precise supervision. However, the training for reward model needs fine-grained annotations, which demands skilled annotators and can be very expensive \cite{DBLP:journals/corr/abs-2305-20050,DBLP:journals/corr/abs-2308-09583}. Additionally, the reward model reflects human preferences, which might introduce bias, and may not always align perfectly with objective correctness or usefulness \cite{wang2024secrets, pitis2023failure}.

\section{Methodology}
\begin{figure}[t]
    \centering
    \resizebox{0.48\textwidth}{!}{
        \subfigure[Train Set]{
            \begin{minipage}[t]{0.45\linewidth}
            \centering
            \includegraphics[width=1\linewidth]{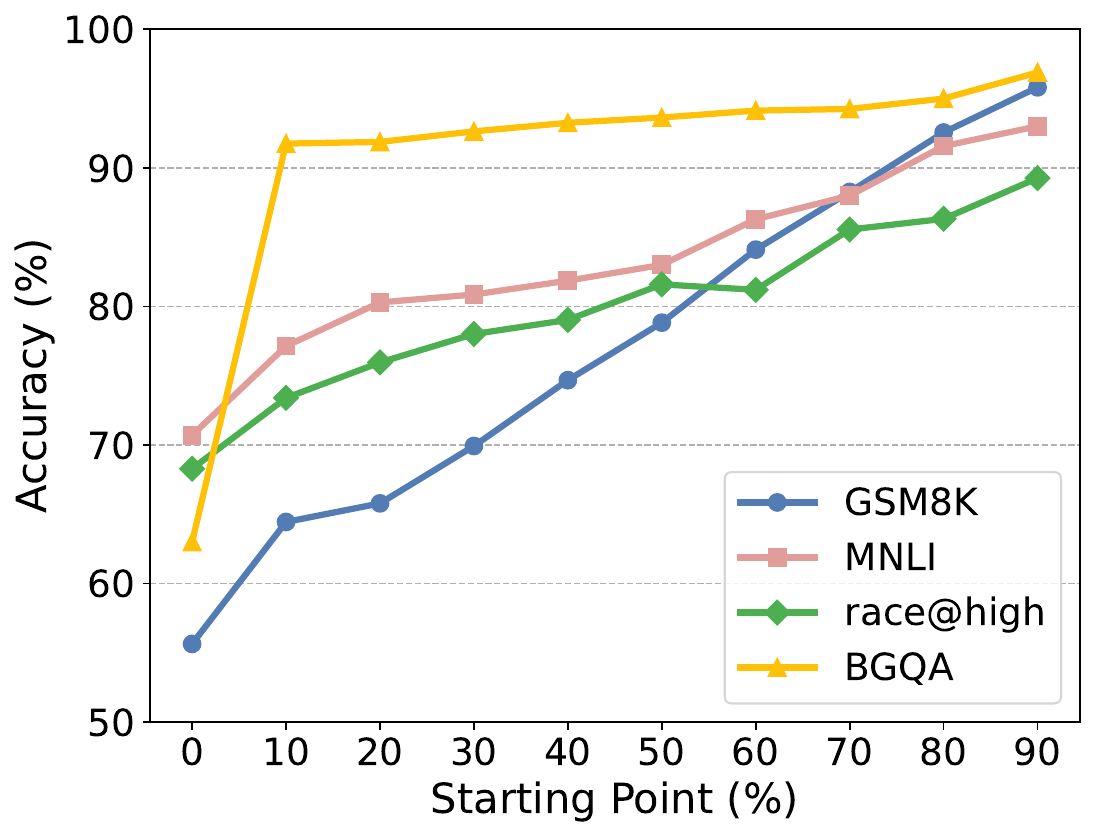}
            \end{minipage}\label{Fig: rollout_percentage_trainset}
        }
        \subfigure[Test Set]{
            \begin{minipage}[t]{0.45\linewidth}
            \centering
            \includegraphics[width=1\linewidth]{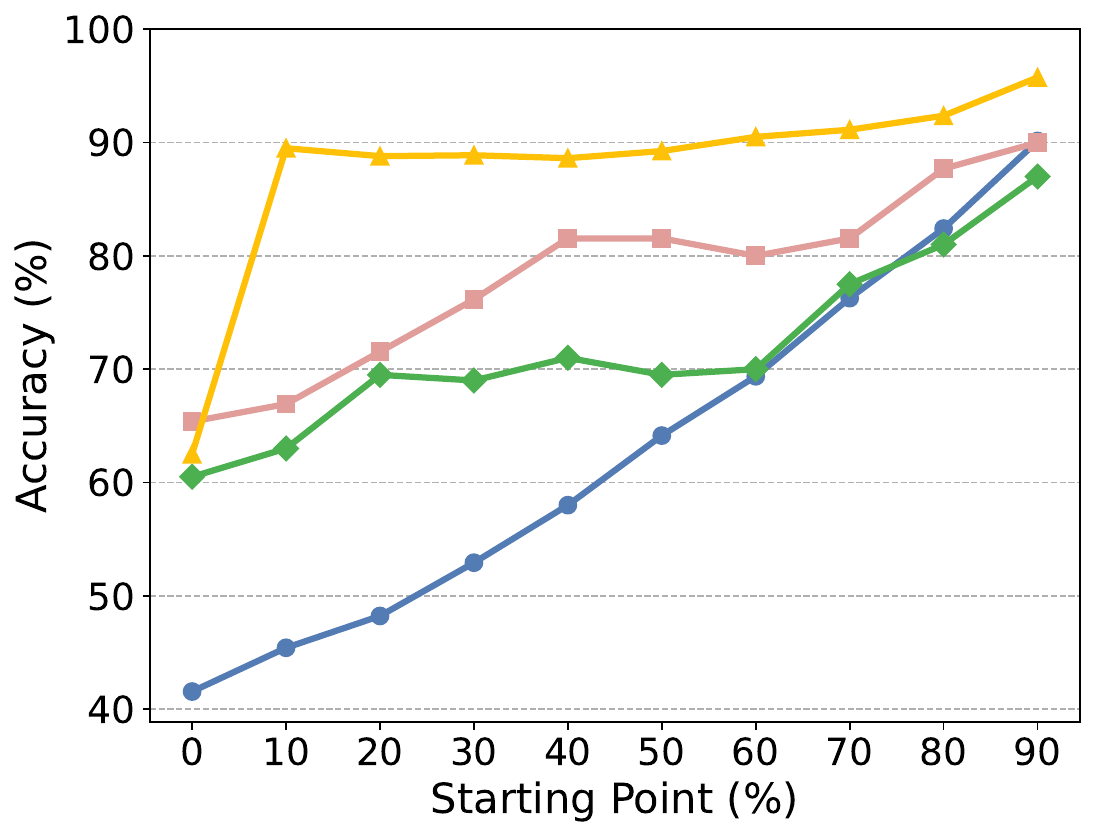}
            \end{minipage}
            \label{Fig: rollout_percentage_testset}
        }
    }
    \vspace{-0.1cm}
    \caption{Accuracy v.s. different start state for exploration. 
    The horizontal axis represents the start state for exploration, with the values indicating the percentage of given actions out of the total actions in the demonstration. 
    The results demonstrate a trend that starting the reasoning from a position closer to the target state makes it easier for the model to obtain a positive reward.
    }
    \label{fig:rollout_percentage}
     \vspace{-0.2cm}
\end{figure}

\begin{figure*}[t]
    \includegraphics[width=\linewidth]{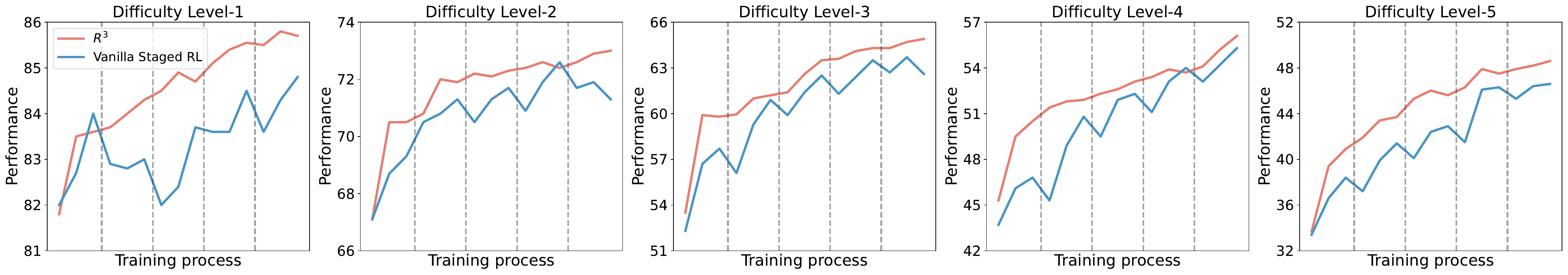}
    \centering
    \vspace{-0.5cm}
 	\caption{
  Learning curves on test sets with $5$ different difficulty level for Staged RL and R$^3$.
  The farther the starting point for exploration is from the target, the higher the difficulty level.
  The horizontal axis represents the training process. The vertical dashed lines indicate the transitions between training stages for staged RL. The experiments are conducted on GSM8K reasoning. Staged RL suffers significant performance drops when transitioning stages, while the performance of R$^3$ improves stably.
          }
	\label{fig:difficulty_test}
 \vspace{-0.15cm}
\end{figure*}

\paragraph{Motivation.}
From our previous analysis in Section \ref{sec:RL with Outcome Supervsion and Process Supervision}, we seek to merge the benefits of outcome and process supervision while avoiding their drawbacks. We aim to develop a method that doesn’t need fine-grained annotations for every step or training a reward model, avoids personal biases by using only golden outcome supervision, and still provides an effect akin to step-level supervision. 
Hence, we assume access only to the outcome-based reward function $rf_o(\cdot)$ and propose \textbf{R}$^3$: Learning \textbf{R}easoning through \textbf{R}everse Curriculum \textbf{R}einforcement Learning.

\subsection{Start Exploration from Intermediate States of Demonstrations}

For a multi-hop reasoning problem, there is a golden answer that can be derived through different reasoning paths. 
We assume to have access to at least one demonstration, i.e., correct reasoning path that leads to a the golden answer as in supervised fine-tuning. 
When the model begins exploration from the initial start state $s_0$, it might face difficulty in obtaining positive rewards as discussed in Section \ref{sec:Outcome Supervision and Process Supervision}. 

Inspired by previous work in the area of RL with demonstrations \cite{DBLP:conf/icml/KakadeL02,DBLP:conf/atal/SubramanianIT16,DBLP:conf/corl/FlorensaHWZA17,openai_1}, we define the set of intermediate states of a given demonstration as $S^{Inter} \subset \mathcal{S}$, and let the policy LM $\pi_\theta$ start exploration from an intermediate state $s_{k} \in \mathcal{S}^{Inter}$ close to the target state: $\pi_\theta (\mathbf{a}_{k+1:T}|{s_k})$ where $\mathbf{a}_{k+1:T}=(a_{k+1},...,a_T)$.
An outcome-based reward function then provides feedback for the final result, serving as a supervisory signal for actions taken after $s_{k}$. 
In this strategy, the trajectory preceding $s_k$ in the demonstration (i.e., $\{s_0,a_1,s_1,...,a_k\}$) can serve as a form of guidance, enabling the model to get positive rewards more easily and avoid getting stuck in directionless, inefficient exploration processes, as shown in Figure \ref{fig:rollout_percentage}.

\subsection{Reverse Curriculum Learning for Step-level Supervision}
Once the policy learns to achieve the goal starting from the selected state close to the target, it can extend its training to more distant states (e.g., $s_{k-1}$), bootstrapping the knowledge it has already acquired. 
At each point, the model faces an easy exploration problem where it is likely to succeed, as it has already learned to solve most of the remaining parts. In this way, a curriculum of gradually increasing exploration difficulty is created, allowing us to provide approximately step-by-step supervisory signals for the model. 
Now the policy gradient can be written as:
\begin{equation}
\begin{aligned} 
  &  \mathbb{E}_{{s_{k}}\sim \mathcal{S}^{Inter}} \Biggl[ \mathbb{E}_{\tau \sim \pi^\mathrm{RL}_\theta(\cdot|s_k)} \Biggl[ \\ 
   &\ \ \ \ \ \ \ \ \ \ \  \sum_{t=k+1}^T  \nabla_\theta \log \pi^\mathrm{RL}_\theta(a_{t}|s_{t-1}) R_o(s_{t-1},a_t) \Biggr] \Biggr] ,
\end{aligned}
\end{equation}
where $\mathcal{S}^{Inter}$ refers to the set of intermediate states of a demonstration sampled from dataset $\mathcal{D}$; $k$ starts from $T-1$ and progressively slides back to $0$. In the final step, the model begins rolling out from the initial state $s_0$, which is equivalent to the original outcome-supervised RL.

In multi-step reasoning, language models may generate a large number of actions (i.e., tokens), making it difficult to enumerate all possible intermediate states and explore from these states. 
Therefore, the number of start states in the reverse curriculum will affect training costs and final reasoning performance.
In our method, we sample $M$ intermediate states from demonstrations either at line breaks (if present) or uniformly, as start states for exploration.
Thus, a reverse curriculum with $M$ stages is created using these selected starting points\footnote{Please note that `stage' here refer to training stages, where the intermediate states sampled in the first stage are those closest to the goal, while the states sampled in the last stage are those farthest from the goal.}. We refer to this method in this paper as \textbf{vanilla staged RL}. 
In our experiments, $M$ is typically $5$ or $6$ and in Section \ref{sec:ablation}, we analyze the impact of the number of stages on reasoning performance.

\subsection{Mixing Start States for Generalization}\label{sec:Mixing Starting Points for Stable Learning}

As shown in preliminary experiments in Figure \ref{fig:difficulty_test}, staged RL may have potential limitations. Models might overfit to simple patterns presented in the early stages of the curriculum and fail to generalize effectively when the difficulty increases,  leading to a degradation of previously acquired knowledge. Furthermore, our findings indicate that staged RL may struggle to adequately capture and model complex interactions and dependencies inherent within the data.
To address this issue, we draw inspiration from the field of multi-task learning \citep{DBLP:journals/corr/Ruder17a,DBLP:journals/tkde/ZhangY22} and treat each stage as an independent task. 
In the final \textbf{R}$^3$, we adopt a mixed strategy to ensure smooth transitions and cooperative optimization between stages of different difficulty levels, stabilizing the training process and enhancing reasoning performance.

\subsection{Reward Design and Policy Optimization}

We employ proximal policy optimization (PPO, \citealp{DBLP:journals/corr/SchulmanWDRK17}) as our basic policy gradient algorithm as it has proved effective in RLHF of LLMs. We apply partial reward ${\epsilon}$ (e.g., $\epsilon = 0.1$) on mathematical reasoning tasks when answer can be extracted and of numeric type to make the reward denser following \cite{DBLP:journals/corr/abs-1709-00103,DBLP:conf/nips/Le0GSH22}:
$$
rf_o(s_{T-1},a_{T}) = \left\{  \begin{aligned} 
1,  \ \ \ \  & \text{answer correct}  \\
\epsilon, \ \ \ \ & \text{answer not correct, but numeric}  \\
0,\ \ \ \    & \text{answer not correct}
\end{aligned}
\right. 
$$
We also design reward functions based on the exploration difficulty, which will be discussed in Section \ref{sec:Difficulty-based Reward Function Design}.
Following \cite{DBLP:journals/ftml/LuRDIOW23}, our total reward is the sum of reward function score and the Kullback-Leibler (KL) divergence between the learned RL policy and initial policy $\pi_\theta^{Init}$ scaled by a coefficient factor $\beta$:
\begin{equation}
\begin{aligned} 
&r_{final}(s_{t-1},a_t) = r_o(s_{t-1},a_t)  \\
&\ \ \ \ \ \ \ \ \ -\beta {\mathrm{KL}\Biggl(}\pi_\theta^{RL} (\cdot|s_{t-1}), \pi_\theta^{Init} (\cdot|s_{t-1})\mathbf{\Biggr)},
\end{aligned}
\end{equation}
We calculate advantages with generalized advantage estimate (GAE) and perform optimization similar to \citet{DBLP:journals/corr/SchulmanWDRK17}. 
Our algorithm is outlined in Algorithm \ref{Algorithm: R3}. We first construct the curriculum datasets of different stages and describes procedures for vanilla staged RL and the final R$^3$.

\begin{table*}[t]
\centering
\caption{Evaluating results on CoT Reasoning. The best results of each dataset is in \textbf{bold} and marked with \underline{underline}, while the second is marked with \underline{underline}. Generally, ``Staged RL'' represents RL with a reverse, staged manner, while R$^3$ represents the final method with mixed stages. While the vanilla staged RL is only slightly better than RL baseline, R$^3$ outperforms all other baselines significantly. }
\resizebox{0.99\textwidth}{!}{ 
\begin{tabular}{lccccccccccc}
\toprule
\multirow{2}{*}{\textsc{\textbf{Method}}} & \multicolumn{2}{c}{\textbf{\textsc{Math Reasoning}}} & \multicolumn{2}{c}{\textbf{\textsc{Logical Reasoning}}} & \multicolumn{2}{c}{\textbf{\textsc{NL Inference}}} & \multicolumn{2}{c}{\textbf{\textsc{Reading Compre.}}} & \multirow{2}{*}{\textsc{\textbf{Average}}} \\
\cmidrule(l){2-9}
 & GSM8K & SVAMP & BGQA$_{\text{main}}$ & BGQA$_{\text{conflict}}$ & MNLI & SNLI & race@High & race@Middle  \\
\cmidrule(l){1-2} 
\cmidrule(l){3-10}

Few-shot  & $15.13$ & $39.62$ & $39.73$ & $34.97$ & $47.69$ & $28.96$ & $38.00$ & $39.20$ & $35.41$ \\
SFT & $41.55$ & $58.40$ & $62.50$ & $57.25$ & $65.38$ & $68.00$ & $60.50$ & $68.00$ & $60.19$ \\
RL  & $44.67$ & $57.30$ & $65.50$ & $58.15$ & $66.15$ & $\underline{69.60}$ & $61.50$ & $69.00$ & $61.48$ \\
Staged RL  & $\underline{47.69}$ & $\underline{61.00}$ & $\underline{67.00}$ & $\underline{58.60}$ & $\underline{67.69}$ & $68.00$ & $\underline{63.00}$ & $\underline{69.50}$ & $\underline{62.81}$ \\
R$^3$  & $\mathbf{\underline{50.49}}$ & $\mathbf{\underline{64.40}}$ & $\mathbf{\underline{67.75}}$ & $\mathbf{\underline{59.35}}$ & $\mathbf{\underline{72.31}}$ & $\mathbf{\underline{72.80}}$ & $\mathbf{\underline{68.50}}$ & $\mathbf{\underline{71.50}}$ & $\mathbf{\underline{65.62}}$ \\
\bottomrule
\end{tabular}
}
\label{table:main results}
\vspace{-1.5mm}
\end{table*}

\begin{table}[ht]
\centering
\caption{Evaluating results of P-CoT reasoning on GSM8K. Our method is marked in \colorbox{lightblue}{blue} and outperforms Few-shot, SFT, and RL. Even against methods needing data augmentation, Codellama + R$^3$ achieves better performance on a 7B model scale.
\textbf{Note} that $\dag$ indicates Tora and Tora-code are trained on additional data in SFT, but this data is not used for R$^3$ as it's not released.
}
\resizebox{0.5\textwidth}{!}{
\begin{tabular}{lccc}
\toprule
 \textbf{\textsc{P-CoT Method}} & \textbf{\textsc{Model Size}} & \textbf{\textsc{Aug Data}} & \textbf{\textsc{Perfor.}} \\ 
\midrule
Glactica + Few-shot & 6.7B &- & $18.6$ \\
Glactica + SFT & 6.7B &- & $57.1$ \\
Glactica + RL & 6.7B &- & $66.1$\\
\blue{Glactica + R\textsuperscript{3}} & \blue{6.7B} &\blue{-} & \blue{$\mathbf{69.3}$} \\
\hdashline
Llama2 + Few-shot & 7B  &- & $18.3$\\
Llama2 + SFT & 7B  &- & $57.7$\\
Llama2 + RL & 7B &- & $63.1$ \\
\blue{Llama2 + R\textsuperscript{3}} & \blue{7B} &\blue{-} & \blue{$\mathbf{68.9}$}\\
\hdashline
Codellama + Few-shot & 7B &- & $32.7$ \\
Codellama + SFT & 7B &- & $63.3$ \\
Codellama + RL & 7B &- & $70.7$\\
\blue{Codellama + R\textsuperscript{3}} &  \blue{7B} &\blue{-} & \blue{$\mathbf{74.2}$}\\
\midrule
\midrule

\multicolumn{4}{c}{\textbf{Models Using Extra Training Data }} \\

MAmmoTH-Coder \cite{DBLP:journals/corr/abs-2309-05653}  & 7B & 260k & $59.4$ \\ 
Tora \cite{DBLP:journals/corr/abs-2309-17452}  & 7B & 16k & $68.8$ \\ 
\blue{Tora \cite{DBLP:journals/corr/abs-2309-17452} + R\textsuperscript{3}}  & \blue{7B} & \blue{16k$^\dag$} & \blue{$\mathbf{73.2}$}   \\ 
Tora-code \cite{DBLP:journals/corr/abs-2309-17452} & 7B & 16k & $72.6$ \\ 
\blue{Tora-code \cite{DBLP:journals/corr/abs-2309-17452} + R\textsuperscript{3}}  & \blue{7B} & \blue{16k$^\dag$} & \blue{$\mathbf{76.3}$}   \\ 

\midrule
\midrule
\multicolumn{4}{c}{\textbf{Larger Models / Close-sourced Models}} \\ 
MAmmoTH-Coder \cite{DBLP:journals/corr/abs-2309-05653} &13B & 260k & $64.7$  \\
MAmmoTH-Coder \cite{DBLP:journals/corr/abs-2309-05653} &34B & 260k & $72.7$  \\
Codex \cite{chen2021evaluating} & N.A. & - & $71.6$ \\
GPT-3.5-Turbo \cite{DBLP:journals/corr/abs-2309-11054} & N.A. & - & $78.0$ \\
GPT-4 \cite{DBLP:journals/corr/abs-2303-08774, DBLP:journals/corr/abs-2308-07921} & N.A. & - & $97.0$ \\ 
\bottomrule
\end{tabular}}

\label{table:P-CoT Results}
\end{table}

\section{Experiments}

\subsection{Experimental Setup}\label{sec: Experimental Setup}
\paragraph{Datasets.}
Given that our work focuses on enhancing the reasoning capabilities of LLMs, we select various task types that require reasoning abilities, including logical reasoning, mathematical reasoning, reading comprehension, and natural language inference (NLI). We also consider program-based reasoning (i.e., P-CoT) for math problem solving following \citet{DBLP:conf/icml/GaoMZ00YCN23}, where we execute the generated Python program to obtain the answer. 

Regarding mathematical reasoning, we choose GSM8K \cite{DBLP:journals/corr/abs-2110-14168} and  SVAMP \cite{DBLP:conf/naacl/PatelBG21}, two widely used datasets. 
For the logical reasoning, we utilize the BoardgameQA (BGQA, \citealp{DBLP:journals/corr/abs-2306-07934}), which is a challenging reasoning task containing contradictory information from various sources. We select its ``main'' subset and ``conflict'' subset. 
For NLI, we select the commonly used datasets SNLI \cite{DBLP:conf/emnlp/BowmanAPM15} and MNLI \cite{DBLP:conf/naacl/WilliamsNB18}, and acquire their rationales from CoT-Collection \cite{DBLP:conf/emnlp/KimJKJYSS23}. 
For reading comprehension, we choose race@Middle and race@High \cite{DBLP:conf/emnlp/LaiXLYH17}, two challenging reading comprehension tasks, and obtain their rationales from CoT-Collection \cite{DBLP:conf/emnlp/KimJKJYSS23}.

\paragraph{Models and baselines.}
For CoT reasoning, we choose Llama2-Base-7B \citep{DBLP:journals/corr/abs-2307-09288} as our backbone model because it is widely used. We include few-shot CoT, SFT and RL as our baselines.
For P-CoT reasoning, we choose Llama2-Base-7B \citep{DBLP:journals/corr/abs-2307-09288}, Glactica \cite{DBLP:journals/corr/abs-2211-09085}, and Codellama-7B \cite{DBLP:journals/corr/abs-2308-12950} as our backbone. We include few-shot P-CoT, SFT and RL as baselines. We also consider recently proposed methods/models that require data augmentation, including MAmmoTH-Coder (7B \& 34B, \citealp{DBLP:journals/corr/abs-2309-05653}), Tora and Tora-coder (7B \& 13B, \citealp{DBLP:journals/corr/abs-2309-17452}).

\paragraph{Implementation details.}
Our training is done with eight A100-80GB GPUs and using DeepSpeed framework \cite{DBLP:conf/kdd/RasleyRRH20}.
For few-shot CoT, we run five times with different demonstrations and report the average performance. For SFT, we set the learning rate to $2e-5$. For each RL-related method, we first perform SFT to warm-up and then perform RL.
We set the partial reward $\epsilon$ to $0.1$ for SVAMP and $0.2$ for GSM8K.
For CoT experiments, we set $\beta$ to $0.05$ in math reasoning and set $\beta$ to $0.3$ in other tasks; for P-CoT experiments, we set $\beta$ to $0.01$.
For mathematical tasks, we perform $50$ epochs for RL and report the best performance, including CoT and P-CoT. For other tasks, we perform $5$ epochs for RL and report the best performance.

\subsection{Experimental Results}\label{sec:Experimental Results}
\paragraph{Results on CoT reasoning.}
The main results are demonstrated in Table \ref{table:main results}. Generally, we can find that: 
(1) RL methods consistently perform better than prompt-based methods and SFT, showing that by continuously performing exploration and learning, models can refine their reasoning capabilities over time, similar to \citep{DBLP:journals/corr/abs-2308-09583}. 
(2) R$^3$ outperforms other baselines in all tasks, with an average improvement of $5.4$ over SFT and $4.1$ over RL, indicating that our method can provide stable and significant optimization.
However, staged RL is only a bit better than the RL baseline, possibly due to overfitting and ineffective stage-to-stage adaptation mentioned before.

Specifically, our method can enhance different reasoning ability of models. For example, on mathematical tasks, R$^3$ shows significant improvements compared to SFT and RL Baselines, suggesting that our method effectively helps models to acquire and refine structured and formal reasoning abilities through exploration. Our method also allows models to handle reasoning tasks with contradictory information (BGQA), demonstrating a notable enhancement in their defeasible reasoning ability (i.e., reasoning with conflicting information guided by preference, \citealp{pollock1987defeasible, DBLP:conf/atal/HechamBC18,DBLP:journals/tplp/MaherTAWC20}).

\paragraph{Results on P-CoT reasoning.}
The evaluating results on program-based reasoning is shown in table \ref{table:P-CoT Results}. We can find that: (1) R$^3$ outperforms other baselines on P-CoT reasoning across all three models. On average, it exceeds SFT by $11.4$ points and surpasses the RL Baseline by $4.2$ points. This demonstrates that our method is not only highly effective but also versatile and adaptable, capable of extending to various reasoning styles like programs.
(2) Compared to other methods that require data augmentation, e.g., MAmmoTH \cite{DBLP:journals/corr/abs-2309-05653}, Tora and Tora-code \cite{DBLP:journals/corr/abs-2309-17452}, Codellama-7B + R$^3$ achieves the better results in 7B-sized models and matches up well with larger models and closed-source model GPT-3.5-Turbo. 
(3) When our method is applied to models like Tora and Tora-code, which were trained with additional data during SFT, it still yields significant performance gain using only the original data in the reinforcement learning phase, demonstrating its adaptability and wide applicability.

\section{Analysis and Discussion}

\begin{figure*}[t]
    \begin{minipage}{\textwidth}
        \begin{minipage}{0.7\textwidth}
            \centering
            \resizebox{0.95\textwidth}{!}{
                \subfigure[Mean Training Reward]{
                    \begin{minipage}[t]{0.30\linewidth}
                    \centering
                    \includegraphics[width=1\linewidth]{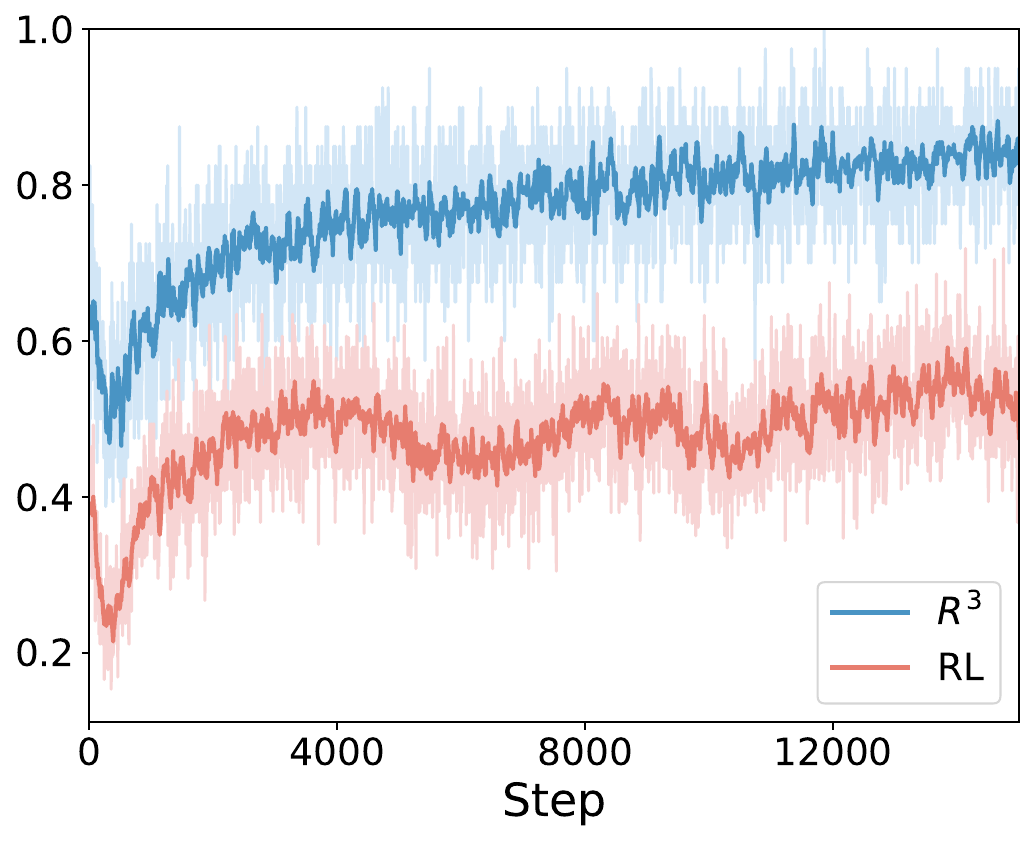}
                    \end{minipage}\label{Fig: gsm8k Mean Training Reward}
                }
                \subfigure[Mean Training Return]{
                    \begin{minipage}[t]{0.31\linewidth}
                    \centering
                    \includegraphics[width=1\linewidth]{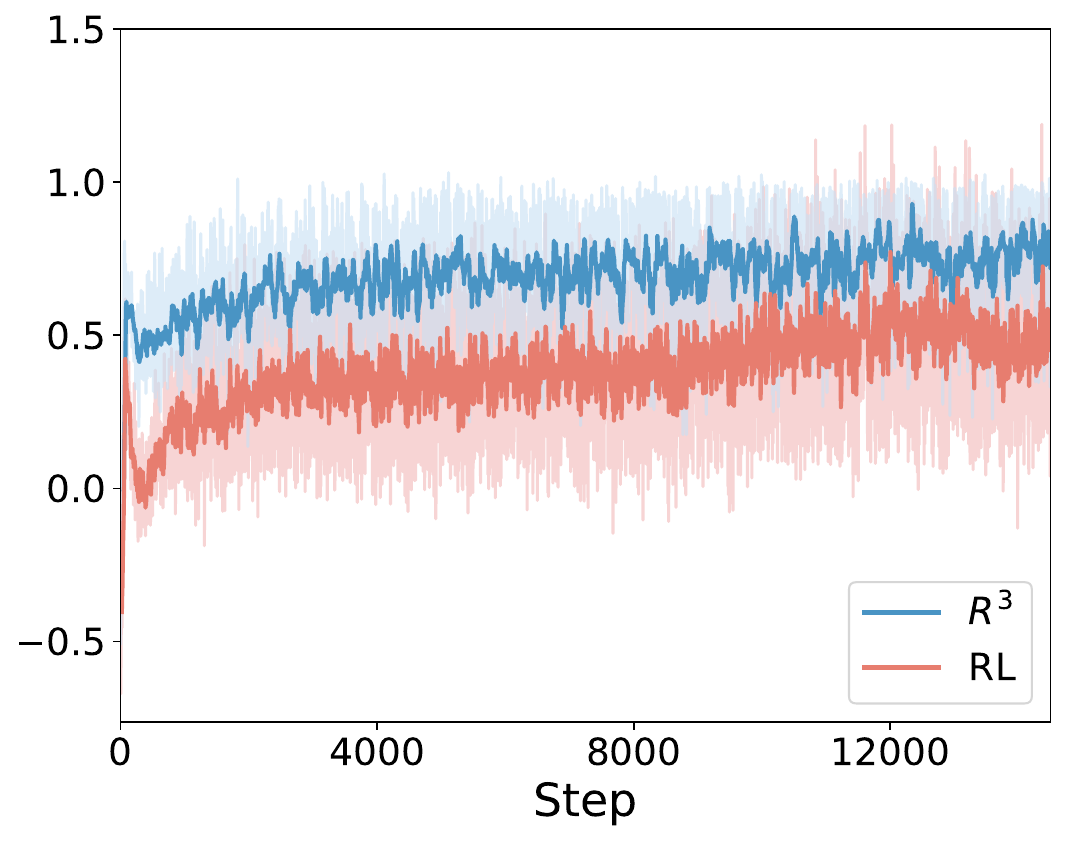}
                    \end{minipage}%
                    \label{Fig: gsm8k Mean Training Return}
                }
                \subfigure[Evaluation Accuracy]{
                    \begin{minipage}[t]{0.3\linewidth}
                    \centering
                    \includegraphics[width=1\linewidth]{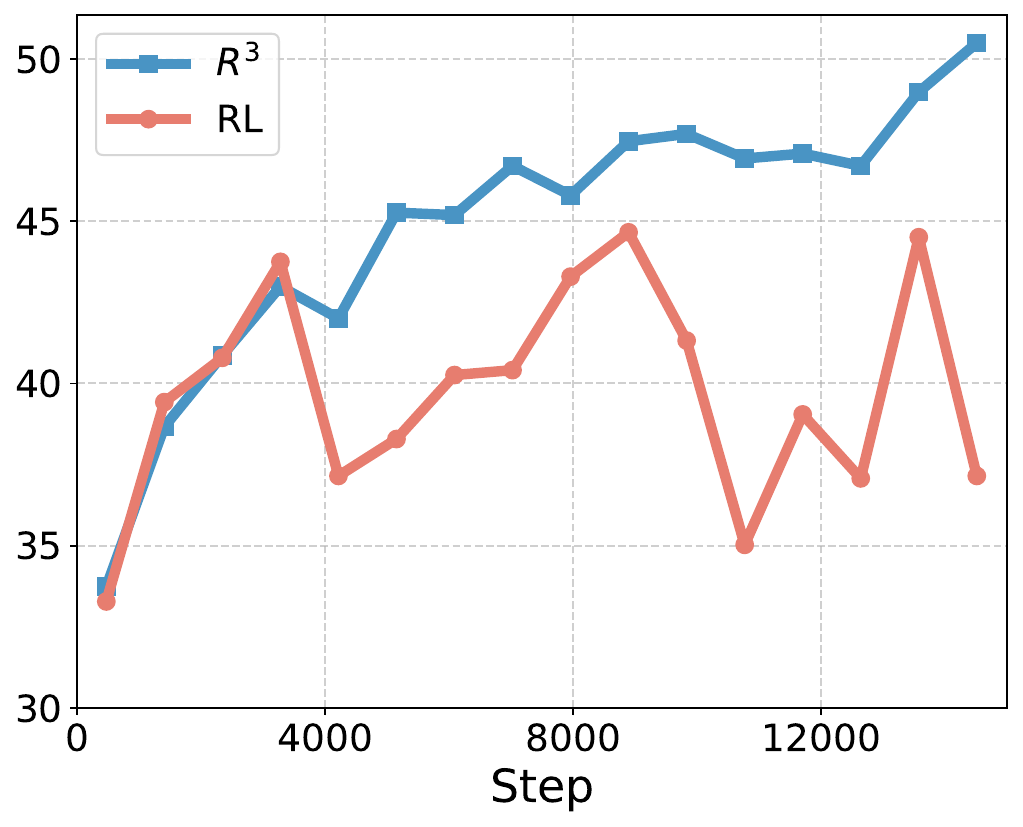}
                    \end{minipage}%
                    \label{Fig: gsm8k Evaluation Accuracy}
                }
            }
            \caption{Training dynamics of RL and R$^3$ on GSM8K CoT, including training reward, training return and evaluation accuracy.}
            \label{fig:training reward_return_acc}
        \end{minipage}
        \begin{minipage}{0.25\textwidth}
            \centering
            \includegraphics[width=0.95\linewidth]{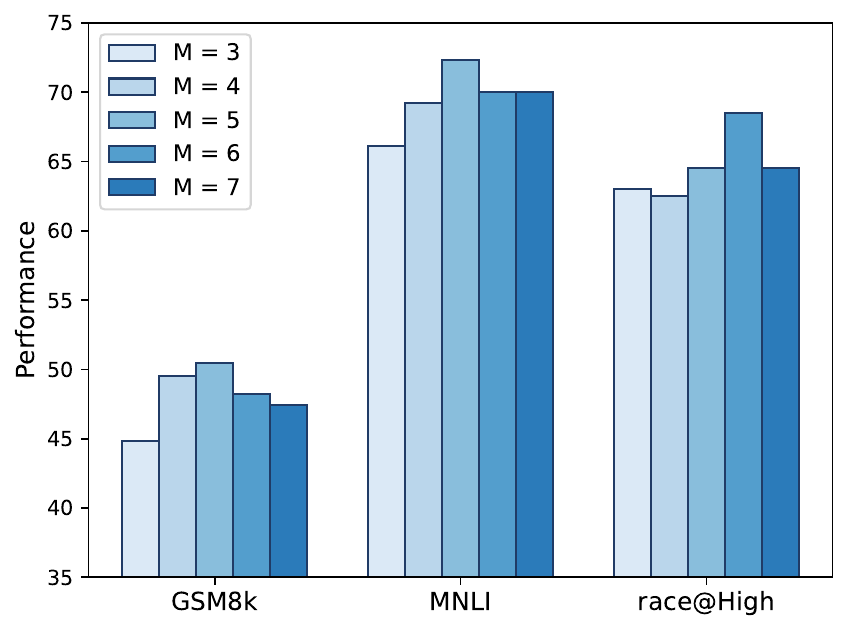}
            \centering
            \caption{Ablation study of different stage numbers $M$. }
            \label{fig:Diffrerent num of stages}
            
        \end{minipage}
    \end{minipage}
\end{figure*}

\subsection{Ablation Study} \label{sec:ablation}

\begin{table}[t]
\centering
\caption{Ablation study on GSM8K CoT, by default $\beta=0.05$, partial reward $\epsilon=0.2$.}
\resizebox{0.4\textwidth}{!}{%
\begin{tabular}{lc}
\toprule
\textbf{Method Setting} & \textbf{Performance} \\ 
\midrule
Llama2-Base 7B + R$^3$ & $\mathbf{50.5}$\\
\hdashline
\ \ \ \   - KL coefficient $\beta=0$ & $46.6$ \\
\ \ \ \  - KL coefficient $\beta=0.1$ & $44.1$ \\
\hdashline
\ \ \ \  - remove partial reward $\epsilon$ & $44.6$ \\
\ \ \ \  - partial reward $\epsilon=0.1$ & $45.9$ \\
\ \ \ \ - partial reward $\epsilon=0.3$ & $45.4$ \\

\midrule
\end{tabular}%
}
\label{tab:ablation}
\vspace{-1.5mm}
\end{table}

\paragraph{KL coefficient $\beta$ and partial reward $\epsilon$.}
We first conduct ablation study on GSM8K CoT to study the impact of $\beta$ and $\epsilon$, and the results are shown in Table \ref{tab:ablation}.\footnote{See Appendix \ref{appendix_ablation_study} for more ablation results on other tasks.}
If we set $\beta=0$, the exploration space of the model becomes unconstrained, and we observe that R$^3$ can still perform well, which is different from the conclusions of previous RL methods where the model may collapse without KL penalty \cite{luong2024reft}. This may be because R$^3$ does not require the model to constantly perform exploration from scratch, reducing the sampling space and making it easier to obtain rewards, thus facilitating training. If we set $\beta=0.1$ to impose higher constraints, we observe a more significant drop in performance, indicating that overly strong KL constraints may hinder the model's optimization.

If we set a small partial reward $\epsilon$ or remove it, R$^3$ obtains a lower performance yet it still outperforms RL and SFT. On the other hand, if we set $\epsilon$ to a bigger value $0.3$, the performance also drops as too large partial reward might lead the model to settle for obtaining simple rewards (outputting numbers) rather than striving for the correct answer.

\paragraph{Number of intermediate states selected $M$.}\label{sec: intermediate states} As mentioned before, if we include all possible intermediate states as starting points, the cost can be extremely high. However, too small value of $M$ might lead to large gaps between stages. Therefore, we need to find a balance and identify an appropriate $M$. We perform ablation experiments and the results in Figure
 \ref{fig:Diffrerent num of stages} show that the performance converges when $M$ reaches an appropriate value, such as $5$ or $6$, and larger $M$ does not yield significant benefits.

\subsection{R$^3$ Delivers Stable Reinforcement Learning}
Figures \ref{Fig: gsm8k Mean Training Reward} and \ref{Fig: gsm8k Mean Training Return} illustrate the training dynamics of vanilla RL and  R$^3$ throughout the training process. We observe that RL encounters instability and fluctuations in training rewards, whereas our method is significantly more stable and yields higher returns. This can be attributed to R$^3$ providing denser, more detailed, and accurate supervisory signals, facilitating model's exploration and learning. The distinction is also evident in test performance, as shown in Figure \ref{Fig: gsm8k Evaluation Accuracy}, where our method achieves more stable improvements. We also provide case studies in Appendix \ref{appendix_case_study} to intuitively show the superiority of our method.
\subsection{Difficulty-based Reward Function Design}\label{sec:Difficulty-based Reward Function Design}
\begin{table}[t]
\centering
\vspace{-2.4mm}
\caption{Performance when adopting different reward functions. The ``Original'' one is the basic reward function that returns $1$ if the answer is correct else $0$. Other functions assign various rewards according to the difficulty of exploration.}
\resizebox{0.49\textwidth}{!}{
\begin{tabular}{lccccc}
\toprule

 \multirow{3}{*}{\textsc{Dataset}} & \multicolumn{5}{c}{\textsc{Reward Function}}   \\

 \cmidrule(l){2-6}

 &Original &$R_{linear}$ &$R_{square}$ &$R_{sqrt}$  &$R_{discount}$ \\
\cmidrule(l){1-1} 
\cmidrule(l){2-6}

 MNLI &$72.3$ &$68.5$  &$68.5$  &$70.0$ &$68.5$\\
 race@High &$68.5$ &$65.0$  &$65.5$  &$66.0$ &$66.0$\\
 GSM8K &$50.5$ &$43.7$  &$41.9$  &$44.1$ &$45.1$\\
 \midrule
  Average &$63.8$ &$59.1$  &$58.7$  &$60.0$ &$59.9$\\
\midrule
\end{tabular}
}
\label{table:different reward functions}
\vspace{-1.5mm}
\end{table}

As mentioned before, when perform exploration from different states of the demonstration, the difficulty for the model to obtain a positive reward varies. This leads to an intuitive question: should we set different amounts of rewards for rollouts of varying difficulty, instead of setting them all to $1$ when the final results are correct? Consequently, we use different variants of the reward function to observe their performance changes. Specifically, assuming the length of a demonstration $\tau$ is $T$: $\tau = (s_0,a_1,s_1,a_2...s_T)$, with the starting point as $s_k$, we approximately define the difficulty of the rolling out process as: $\mu = (T-k)/T$.

We then consider different reward functions related to the difficulty. These functions have different trends of change on the slope, including linear reward function $R_{linear} = \mu$, square reward function $R_{square} = \mu^2$, and square root reward function $R_{sqrt} = \sqrt{\mu}$.
Inspired by the conception of discount factor in RL, we also consider another discount reward function: $R_{discount} = \gamma^{(T-k)}$, where $\gamma=0.9$.
Experiments in Table \ref{table:different reward functions} show counter-intuitive results that these modified reward functions do not bring performance improvements, but rather, performance decreases. This implies that we should treat each start state fairly.
\subsection{Analysis of Training Data Construction}
\begin{figure}[t]
    \centering
    \resizebox{0.48\textwidth}{!}{
        \subfigure[Impact of Data Scale]{
            \begin{minipage}[t]{0.47\linewidth}
            \centering
            \includegraphics[width=1.05\linewidth]{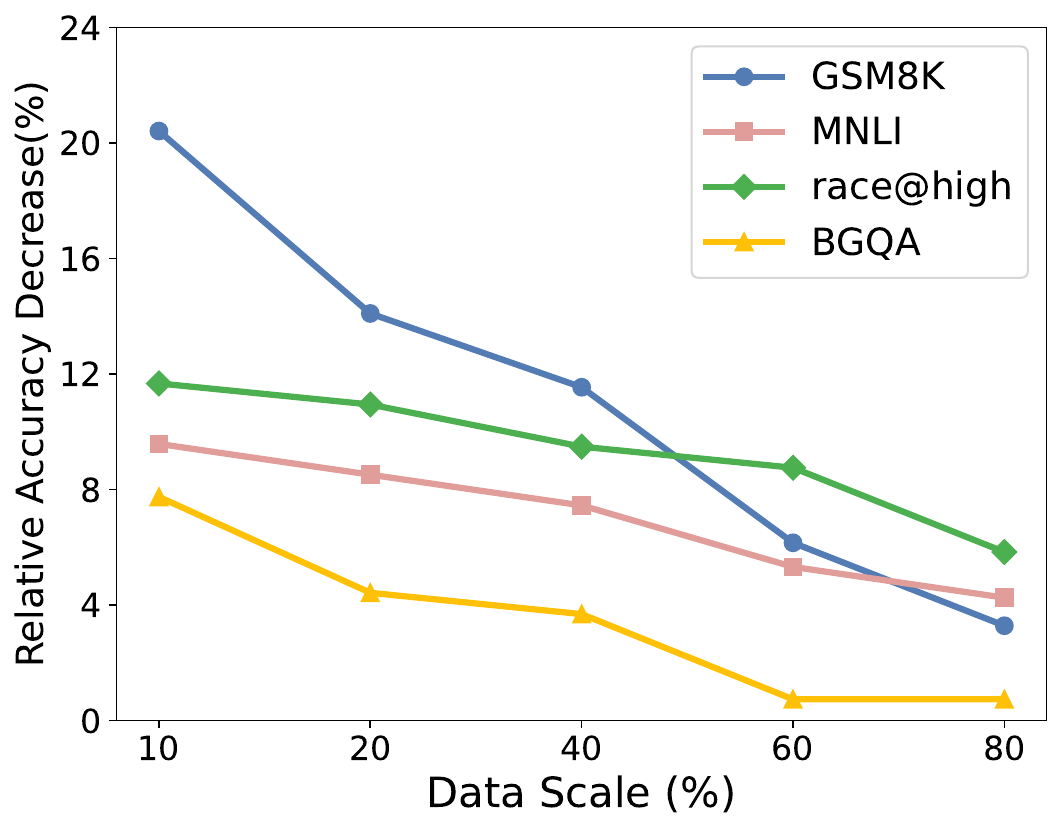}
            \end{minipage}\label{fig: scaling_diff}
        }
        \subfigure[Impact of Data Composition]{
            \begin{minipage}[t]{0.47\linewidth}
            \centering
            \includegraphics[width=1.1\linewidth]{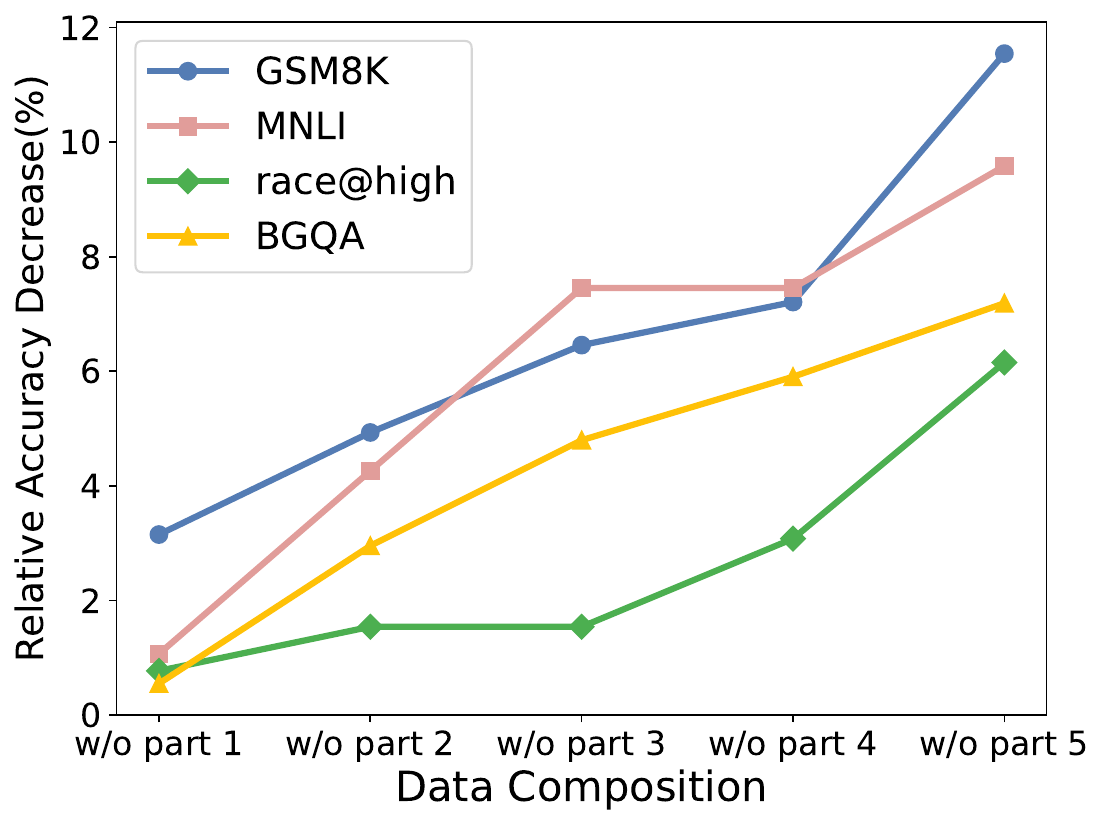}
            \end{minipage}
            \label{fig: mix_part_diff}
        }
    }
    \caption{
    Impact of data scale and composition. The vertical axis represents the percentage of performance decrease relative to training with full dataset. The horizontal axis of the left subfigure represents the amount of data used, while the horizontal axis of the right subfigure, labeled ``w/o part $j$'', indicates removing a part of training data corresponding to a specific difficulty level $j$. 
    }
    \vspace{-1.5mm}
    \label{fig:scaling}
\end{figure}
\paragraph{Scaling of training data.}\label{para: label scaling}
We first study the data efficiency of R$^3$, and the results are shown in Figure \ref{fig: scaling_diff}.
Overall, as the amount of data decreases, the performance of R$^3$ shows a decreasing trend. However, the sensitivity of R$^3$ to data scale varies by task. For instance, on GSM8K, using a limited amount of data leads to a significant decline in performance. This may be because such tasks require a large amount of data to learn enough specialized mathematical knowledge to enable the model to generalize. In contrast, for BGQA, even with limited data scale, the model might still achieve better generalization performance by learning patterns and relationships in the language. 
Moreover, we demonstrate the absolute values of performance in Appendix \ref{appendix: data scale}, and the results show that R$^3$ can outperform the RL baseline with only a portion of the data.

\paragraph{Which part of data matters?}
Next we investigate which part of training data is crucial. We remove training data of varying difficulties (i.e., the farther the starting point is from the target, the greater the difficulty) and conduct experiments. Results in Figure \ref{fig: mix_part_diff} demonstrate a trend that removing the more difficult data results in poorer performance, highlighting the importance of challenging data. Conversely, removing the simplest data does not significantly degrade performance. We also provide the absolute performance values in the Appendix \ref{appendix: data parts}.

\section{Related Work}
\paragraph{Reasoning with large language models.}
Multi-hop complex reasoning is considered one of the most challenging task for LLMs \citep{DBLP:journals/corr/abs-2112-11446,DBLP:journals/corr/abs-2108-07258,DBLP:journals/corr/abs-2212-09597}, and researchers have developed several categories of methods, including prompting, supervised fine-tuning methods and reinforcement learning methods. 
Prompting, with chain-of-thought as a representative one, involves constructing demonstrations and instructions in the prompt to improve model's reasoning performance \cite{DBLP:conf/nips/Wei0SBIXCLZ22,DBLP:conf/nips/KojimaGRMI22,DBLP:conf/emnlp/XiJZZGLGZH23,DBLP:journals/corr/abs-2309-15402}. However, they proved to be sensitive to many factors and model-dependent \cite{DBLP:journals/corr/abs-2302-00093,DBLP:conf/emnlp/ZellersBSC18,DBLP:journals/corr/abs-2205-03401}. 
In SFT, models are trained with collected rationales, and their effectiveness largely relies on the scale and quality of the training data \cite{DBLP:journals/corr/abs-2308-01825,DBLP:journals/corr/abs-2309-12284,DBLP:journals/corr/abs-2309-05653}, necessitating considerable effort in gathering annotations. 
RL is also used in LLM reasoning, which will be discussed in detail in the next paragraph.

\paragraph{Reinforcement learning for large language models.}
RL has garnered much attention in LLM alignment \cite{DBLP:journals/corr/abs-2112-00861,DBLP:journals/corr/abs-2204-05862,DBLP:conf/nips/Ouyang0JAWMZASR22,DBLP:journals/corr/abs-2307-04964,DBLP:journals/corr/abs-2401-06080}, and has been applied in many other tasks like summarization \cite{DBLP:conf/nips/Ouyang0JAWMZASR22,DBLP:journals/corr/abs-2009-01325}, web navigation \cite{DBLP:journals/corr/abs-2112-09332,DBLP:conf/acl/QinCJYLZLHDWXQL23} and machine translation \cite{DBLP:journals/corr/abs-2308-08998}. 
There are also some work explores enhancing model's reasoning capabilities with RL, based on outcome supervision or process supervision \cite{DBLP:journals/corr/abs-2305-20050,DBLP:journals/corr/abs-2308-09583,DBLP:journals/corr/abs-2309-02144,luong2024reft}. Furthermore, these two types of supervision are also utilized to perform answer reranking at inference time \cite{DBLP:journals/corr/abs-2211-14275,DBLP:journals/corr/abs-2110-14168,DBLP:journals/corr/abs-2311-09724}, which involves training a reward model based on either outcome or process supervision to rank multiple generated solutions and select the top one. These approaches are orthogonal to our method and can be seamlessly integrated for further improvement. 

\paragraph{Reinforcement learning with reverse curriculum.}
In goal-oriented RL, reverse curriculum learning \cite{florensa_automatic_2018,florensa_reverse_2018}
effectively addresses the problem of sparse rewards \cite{ladosz_exploration_2022}. 
This method involves initially training the agent achieve the target from a starting point near the target, and subsequently relocating the starting point to more distant positions \cite{10.1109/IROS51168.2021.9636842}.
Notably, methods that sample starting points from intermediate states of quality demonstrations \cite{subramanian_exploration_2016,popov_data-efficient_2017} and trajectories are commonly applied to tasks like the games 
 \cite{hosu2016playing,openai_1} and robotics \cite{peng_deepmimic_2018,nair_overcoming_2018,plappert2018multigoal}. 
We employ such strategy to address the issue of sparse rewards in outcome supervision of LLM reasoning and provide an effect akin to process supervision.

\section{Conclusion and Future Work}
In this work, we rethink the existing supervision paradigms of reinforcement learning for large language model reasoning, and propose R$^3$ that employs only outcome supervision to achieve the benefits of process supervision via reverse curriculum reinforcement learning. 
We perform thorough experiments on natural language-based and program-based CoT to demonstrate the effectiveness of our method. Moreover, we conduct detailed ablation and analysis to showcase the stability and operating mechanism of our method.
In the future, we will attempt to scale up the model size for better performance. Additionally, we will explore the impact of training data with larger scale and diversity on R$^3$.

\section*{Impact Statements}
This paper presents work whose goal is to advance the field of Machine Learning. There are many potential societal consequences of our work, none which we feel must be specifically highlighted here.

\bibliography{example_paper}
\bibliographystyle{icml2024}

\newpage
\appendix
\onecolumn
\section{Algorithm}\label{appendix_algo}

\begin{algorithm}[ht]
\caption{R$^3$}
\label{Algorithm: R3}
  \SetKwData{Left}{left}\SetKwData{This}{this}\SetKwData{Up}{up}
  \SetKwFunction{Union}{Union}\SetKwFunction{FindCompress}{FindCompress}
  \SetKwInOut{Input}{Input}\SetKwInOut{Output}{Output}
   \SetKwProg{myproc}{Procedure}{}{}
   \KwIn{Policy language model $\pi_\theta$, training data $\mathcal{D}$ with $N$ data points, maximum rollout length $T$, number of stages $M$, outcome-based reward function $rf_o
   (\cdot)$.}
   ]
   \vspace{3pt}
   Initialize policy model $\pi_\theta$;
   
   \myproc{{{\textnormal{Construct reverse curriculum datasets:}}}}{
        $\mathcal{D}_{list}  \gets [\ ]$;
        
        \For{\textnormal{stage} $ m \gets 1...M$}{
            $\mathcal{D}^m  \gets \emptyset$;

            \For{\textnormal{data point} $(s_0,\mathbf{a})=\{s_0,a_1,...,s_{T-1},a_T\}$ \textnormal{in} $\mathcal{D}$ }{
                Select an intermediate state $s_k$ as the start state for stage $m$;\\
                $\mathcal{D}^m  \gets \mathcal{D}^m \cup (s_k,\mathbf{a}_{k+1:T})$;
            }
       }
       $\mathcal{D}_{list} . \text{append}(\mathcal{D}^m)$ 
   }
   \myproc{{{\textnormal{Rinforcement learning in a reverse, staged manner:}}}}{      
        \For{\ \ $\mathcal{D}^m$\ \  \textnormal{in}\ \  $\mathcal{D}_{list}$\ \ }{
            Perform Reinforcement Learning with $\pi_\theta$ and $rf_o(\cdot)$ on $\mathcal{D}^m$;
        }
    }
    \myproc{{{ \textnormal{Rinforcement learning with mixed stages:}}}}{
        
        $\mathcal{D}^{mixed} \gets \bigcup\limits_{m=1}^{M}  \mathcal{D}_{list}[m] $;

        Perform Reinforcement Learning with $\pi_\theta$ and $rf_o( \cdot )$ on $\mathcal{D}^{mixed}$;
    
    }
\end{algorithm}

\section{Additional Experiments}
\subsection{Ablation Study}\label{appendix_ablation_study}

In Table \ref{tab:appendix_ablation}, we conduct supplementary ablation studies on Section \ref{sec:ablation}, providing results on BGQA$_{\text{main}}$, MNLI and race@High datasets. 
We can observe that if we set $\beta=0.4$, imposing a stronger KL constraint, there will be a noticeable decrease in performance. If we set $\beta$ to $0$ or $0.1$, the performance loss is not as pronounced but still falls below the optimal result.

\begin{table}[ht]
\centering
\caption{Ablation study on BGQA$_{\text{main}}$, MNLI and race@High, by default $\beta=0.3$.}
\resizebox{0.38\textwidth}{!}{%
\begin{tabular}{lc}
\toprule
\textbf{Dataset} & \textbf{Performance} \\ 
\midrule
BGQA$_{\text{main}}$ & $\mathbf{67.8}$\\
\ \ \ \  - KL coefficient $\beta=0$ & $66.3$ \\
\ \ \ \  - KL coefficient $\beta=0.2$ & $66.5$ \\
\ \ \ \  - KL coefficient $\beta=0.4$ & $65.5$ \\
\hdashline
MNLI & $\mathbf{72.3}$\\
\ \ \ \   - KL coefficient $\beta=0$ & $70.0$ \\
\ \ \ \   - KL coefficient $\beta=0.2$ & $70.0$ \\
\ \ \ \   - KL coefficient $\beta=0.4$ & $68.4$ \\
\hdashline
race@High & $\mathbf{68.5}$\\
\ \ \ \   - KL coefficient $\beta=0$ & $64.5$ \\
\ \ \ \   - KL coefficient $\beta=0.2$ & $66.5$ \\
\ \ \ \   - KL coefficient $\beta=0.4$ & $64.5$ \\
\midrule
\end{tabular}%
}
\label{tab:appendix_ablation}
\end{table}

\subsection{Experimental Results of Data Scale}\label{appendix: data scale}
As a supplement to Section \ref{para: label scaling}, Table \ref{tab: scaling} presents detailed values of performance. The table illustrates that R$^3$ achieves performance comparable to full-data training of SFT and RL baselines, using only a fraction of the available data.

\begin{table}[htbp]
    \centering
    \caption{Impact of data scale}
    \begin{tabular}{ccccccccc}
        \toprule
        \multirow{2}{*}{Dataset} & \multicolumn{6}{c}{R$^3$ with Data Scaling (\%)} & \multicolumn{2}{c}{Baseline (Full Train Set)} \\
        \cmidrule(lr){2-7} \cmidrule(lr){8-9}
         & 10 & 20 & 40 & 60 & 80 & 100 & SFT & RL \\
        \midrule
        \multirow{1}{*}{GSM8K} & 40.2 & 43.4 & 44.7 & 47.4 & 48.8 & 50.5 & 41.6 & 44.7 \\
        \multirow{1}{*}{MNLI} & 65.4 & 66.2 & 66.9 & 68.5 & 69.2 & 72.3 & 65.4 & 66.2 \\
        \multirow{1}{*}{race@High} & 60.5 & 61.0 & 62.0 & 62.5 & 64.5 & 68.5 & 60.5 & 61.5 \\
        \multirow{1}{*}{BGQA} & 62.5 & 64.8 & 65.3 & 67.3 & 67.3 & 67.8 & 62.5 & 65.5 \\
        \bottomrule
    \end{tabular}
    \label{tab: scaling}
\end{table}

\subsection{Impact of different parts of data.}\label{appendix: data parts}

Table \ref{tab: mix parts} and Table \ref{tab: mix parts raceHigh} present the accuracy achieved when training the model without specific data parts. Notably, columns 1 through 5 ( For race@High, columns1 through 6 ) signify the ascending difficulty levels of excluded training data, with higher part numbers indicating greater difficulty. The ``All Parts'' column reflects accuracy when utilizing the entire dataset. Furthermore, based on the results in Section \ref{sec: intermediate states}, we can conclude that for the race@High dataset, optimal performance can be achieved when the number of intermediate states $M$ is set to $6$. Therefore, we supplement experiments with race@High containing $6$ data parts in Table \ref{tab: mix parts raceHigh}.

\begin{table}[h!]
\centering
\caption{Comparison of accuracy in training on different data parts}
\begin{tabular}{lcccccccc}
\toprule
\multirow{2}{*}{Dataset} & \multicolumn{5}{c}{w/o Part} & \multirow{2}{*}{All Parts} & \multicolumn{2}{c}{Baseline (Full Train Set)} \\
\cmidrule(lr){2-6} \cmidrule(lr){8-9}
 & 1 & 2 & 3 & 4 & 5 & & SFT & RL \\
\midrule
GSM8K & 48.9 & 48.0 & 47.2 & 46.9 & 44.7 & 50.5 & 41.6 & 44.7 \\
MNLI & 71.5 & 69.2 & 66.9 & 66.9 & 65.4 & 72.3 & 65.4 & 66.2 \\
race@High & 64.5 & 64.0 & 64.0 & 63.0 & 61.0 & 65.0 & 60.5 & 61.5 \\
BGQA & 67.4 & 65.8 & 64.5 & 63.8 & 62.9 & 67.8 & 62.5 & 65.5 \\
\bottomrule
\end{tabular}
\label{tab: mix parts}
\end{table}

\begin{table}[h!]
\centering
\caption{Performance for race@High with $6$ intermediate states}
\begin{tabular}{lccccccccc}
\toprule
\multirow{2}{*}{Dataset} & \multicolumn{6}{c}{w/o Part} & \multirow{2}{*}{All Parts} & \multicolumn{2}{c}{Baseline (Full Train Set)} \\
\cmidrule(lr){2-7} \cmidrule(lr){9-10}
 & 1 & 2 & 3 & 4 & 5 & 6 & & SFT & RL \\
\midrule
race@High & 65.5 & 63.5 & 63.5 & 63.0 & 61.0 & 62.0 & 68.5 & 60.5 & 61.5 \\

\bottomrule
\end{tabular}
\label{tab: mix parts raceHigh}
\end{table}

\section{Prompts}
We follow the Alpaca \cite{alpaca} prompts format in our experiments. The specific prompts are as follows.

\lstset{
  basicstyle=\footnotesize\ttfamily,
  columns=fullflexible,
  breaklines=true,
  frame=lines,
  extendedchars=true,
  escapechar=@,
  literate={á}{{\'a}}1 {ã}{{\~a}}1 {é}{{\'e}}1 {£}{{\pounds}}1 {–}{{-}}1 {’}{{'}}1,
}
\begin{lstlisting}[caption={Prompts used in R$^3$ experiments}]
Below is an instruction that describes a task. 
Write a response that appropriately completes the request.

### Instruction:
{instruction}

### Response:
\end{lstlisting}

\section{Case Study}\label{appendix_case_study}
We provide case studies of R$^3$ and vanilla RL on GSM8K-CoT, GSM8K-P-CoT and MNLI Datasets. Wrong reasoning steps are highlighted in \textcolor{red}{\textbf{red}}, and reasoning steps corrected by the R$^3$ method are indicated in \textcolor{green}{\textbf{green}}. It is evident that the model trained by R$^3$ has clearer logic and more accurate reasoning when facing complex reasoning tasks, often achieving better task completion.

\begin{figure}[htbp]
    \includegraphics[width=0.96\linewidth]{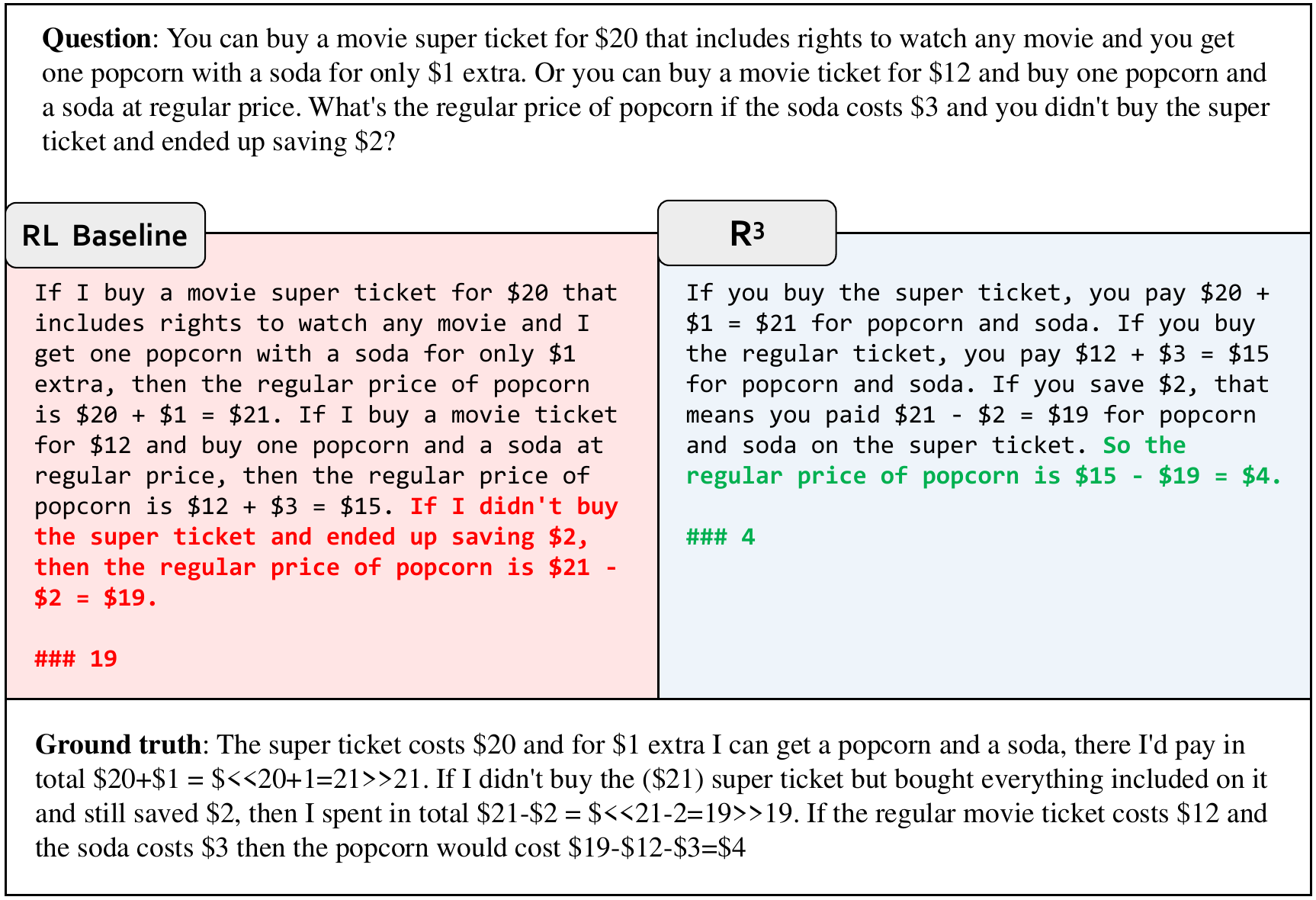}
    \centering
 	\caption{Comparison of RL Baseline and \textbf{R}$^3$ on GSM8K-CoT.}
	\label{fig:case_study_gsm8k_cot}
\end{figure}

\begin{figure}[htbp]
    \includegraphics[width=0.96\linewidth]{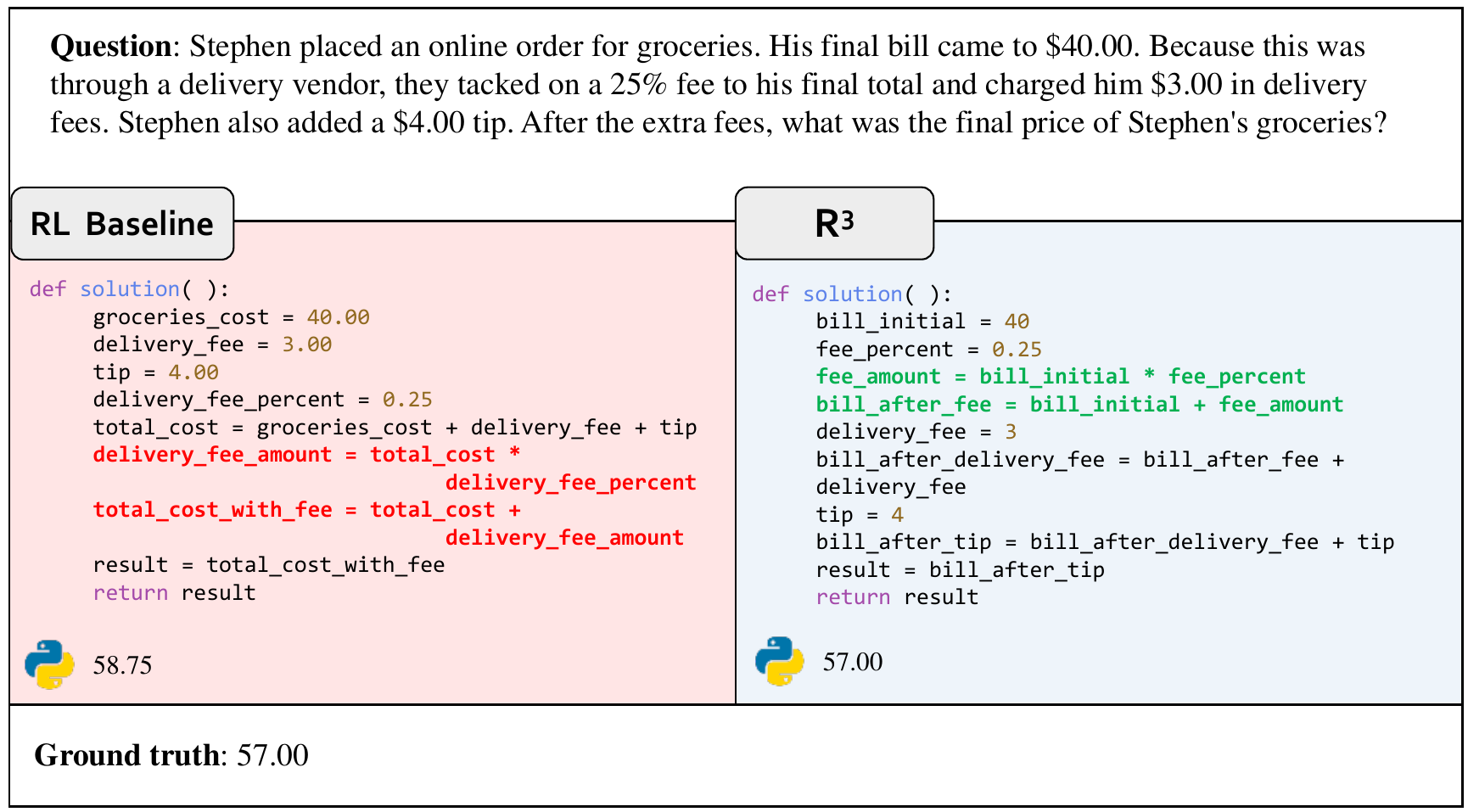}
    \centering
 	\caption{Comparison of RL Baseline and \textbf{R}$^3$ on GSM8K-P-CoT.}
	\label{fig:case_study_gsm8k_pot}
\end{figure}

\begin{figure}[htbp]
    \includegraphics[width=0.96\linewidth]{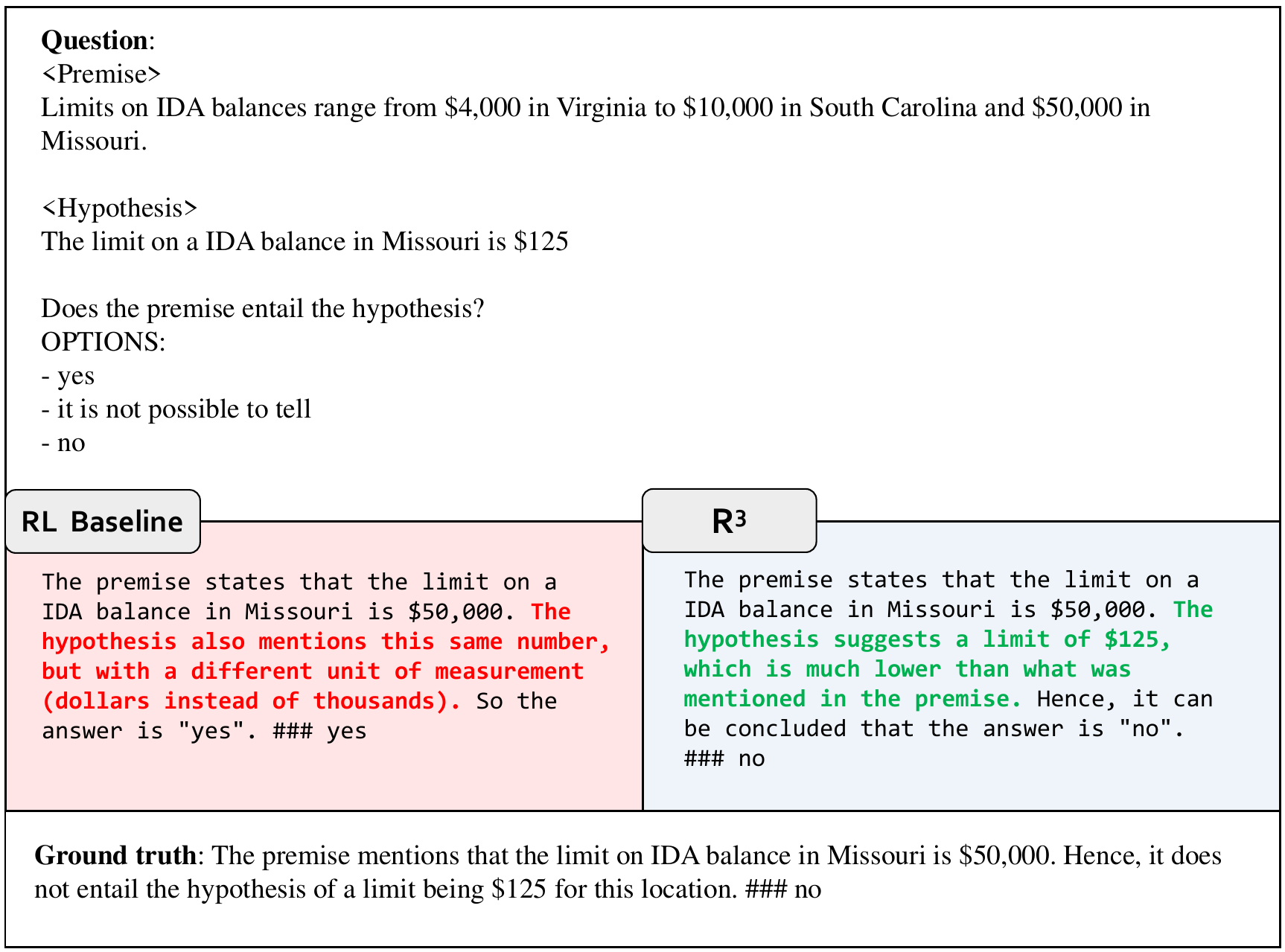}
    \centering
 	\caption{Comparison of RL Baseline and \textbf{R}$^3$ on MNLI.}
	\label{fig:case_study_mnli}
\end{figure}


\end{document}